\documentclass{IOS-Book-Article}     

\usepackage[utf8]{inputenc}
\usepackage[pdftex,bookmarks,colorlinks=false]{hyperref}
\usepackage{graphicx}
\usepackage{filecontents}
\usepackage{setspace}

\begin{document}
\begin{frontmatter}          

\title{Application of Topic Models to Judgments from
       Public Procurement Domain}
\runningtitle{Application of Topic Modelling to Judgments from
       Public Procurement Domain}

\author{\fnms{Michał} \snm{Łopuszyński}}
\address{Interdisciplinary Centre for Mathematical and Computational
Modelling, University of Warsaw, Pawińskiego 5a, 02-106 Warsaw, Poland}
\runningauthor{M. Łopuszyński}
\end{frontmatter}

{\scriptsize
    This a draft version of the paper published in ``Legal Knowledge an
  Information Systems, JURIX 2014: The Twenty-Seventh Annual Conference'',
  Frontiers in Artificial Intelligence and Applications, Volume 271, edited
  by Rinke Hoekstra, IOSPress, 2014.  The final publication is available
  from \href{http://dx.doi.org/10.3233/978-1-61499-468-8-131}{IOSPress}.
  }
\smallskip

Topic modelling algorithms are statistical methods capable of
detecting common themes present in the analyzed text corpora.
In this work, the latent Dirichlet allocation (LDA) is used
~\cite{Blei2003}.
It operates on documents in the bag-of-words representation and returns a
set of detected topics (i.e., groups of words and their probabilities
reflecting the importance in particular topic). In addition, the topic
proportions for each document can be determined. In this paper, I
demonstrate the utility of the LDA method for analyzing judgments from public
procurement field. In addition, I propose combining LDA with a keyphrase
extraction scheme, as it improves topics interpretability and computational
performance.

For this study, a corpus of 13 thousand judgments of the National Appeal
Chamber was used. National Appeal Chamber is competent for the examination
of the appeals lodged in the contract award procedures in Poland. The
judgments covered the period between 12.2007 and 05.2014.  For LDA, the
Gibbs sampling method implemented in the MALLET package
(\url{http://mallet.cs.umass.edu/}) was applied. After preliminary tests,
the number of topics was adjusted to 20. Lower values yield many overly
broad topics, and higher values increased granularity and provided many
topics without a clear interpretation.

\begin{figure}[!b]
\centering
\includegraphics[width=0.9\textwidth]{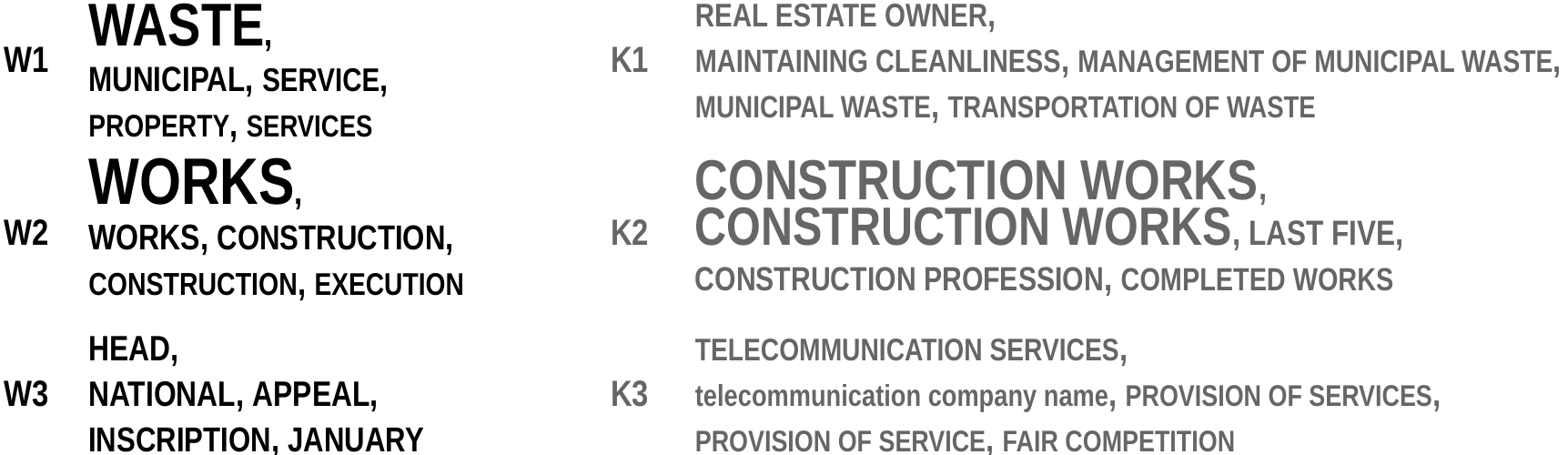}
\caption{
         ''Word clouds'' for sample themes generated by the LDA method.
          W1--W3 were generated using bag-of-words, whereas K1--K3
          resulted from LDA on keyphrases.
         \label{fig:WordClouds}
         }
\end{figure}
Sample themes detected by the LDA method within the analyzed corpus of
judgments are presented in Figure~\ref{fig:WordClouds}.  Initially, the LDA
procedure was performed on the bag-of-words representation.  The obtained
themes looked promising. E.g., topic W1 is clearly related to the municipal
waste management, W2 deals with construction works.  Nevertheless, many
topics seemed difficult to interpret, such as, e.g., topic W3.  Therefore,
the LDA algorithm was ran also on the reduced representation of documents
consisting of auto-detected keyphrases, which were extracted using
unsupervised approach (see \cite{Jungiewicz2014} and references
therein). To a large extent, it was possible to match the topics generated
from plain words with the keyphrase case (compare K1 and W1 both related to
waste management or K2 and W2 which deal with construction works). However,
the keyphrase method detected larger number of clearly interpretable
topics. See, e.g.,  K3 related to telecommunication services, which was not
present among the plain word topics.  Another practical advantage is that
the keyword method is more computationally efficient, because of the
more compact document representation. Full run on the precalculated keyword
representation took 12 minutes, whereas similar analysis
on words lasted over 2 hours on the same 4 CPU cores.

\begin{figure}[!ht]
\centering
\includegraphics[width=0.49\textwidth]{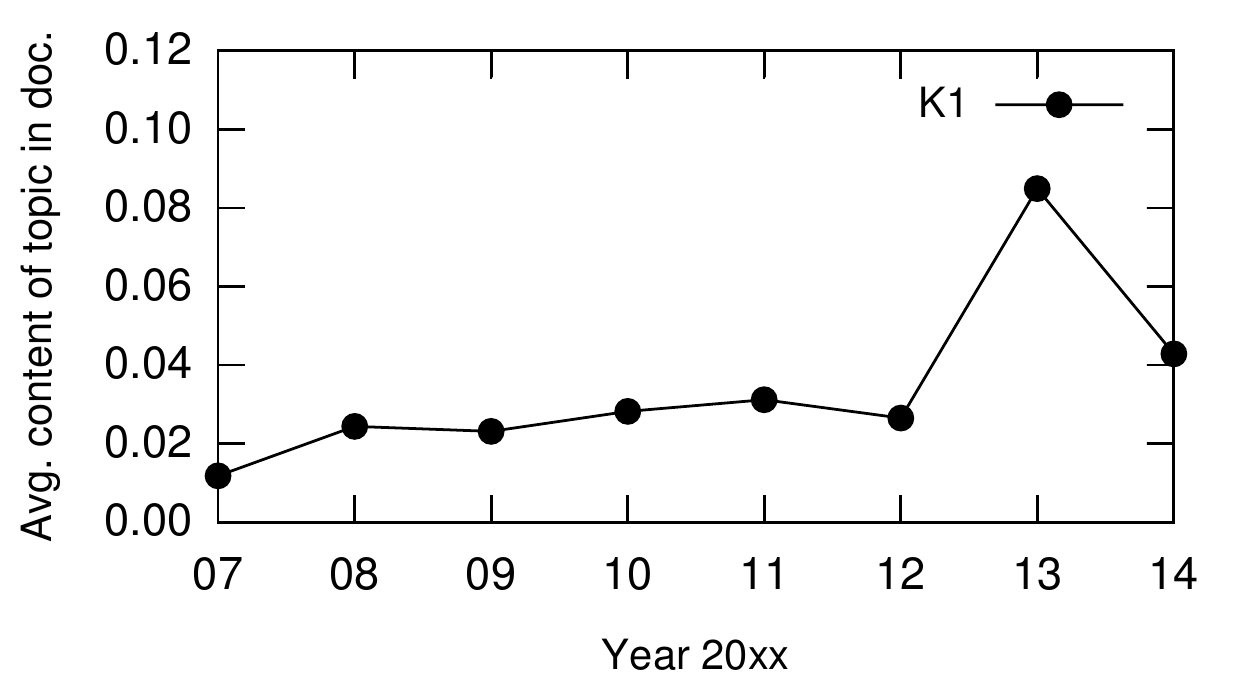}
\includegraphics[width=0.49\textwidth]{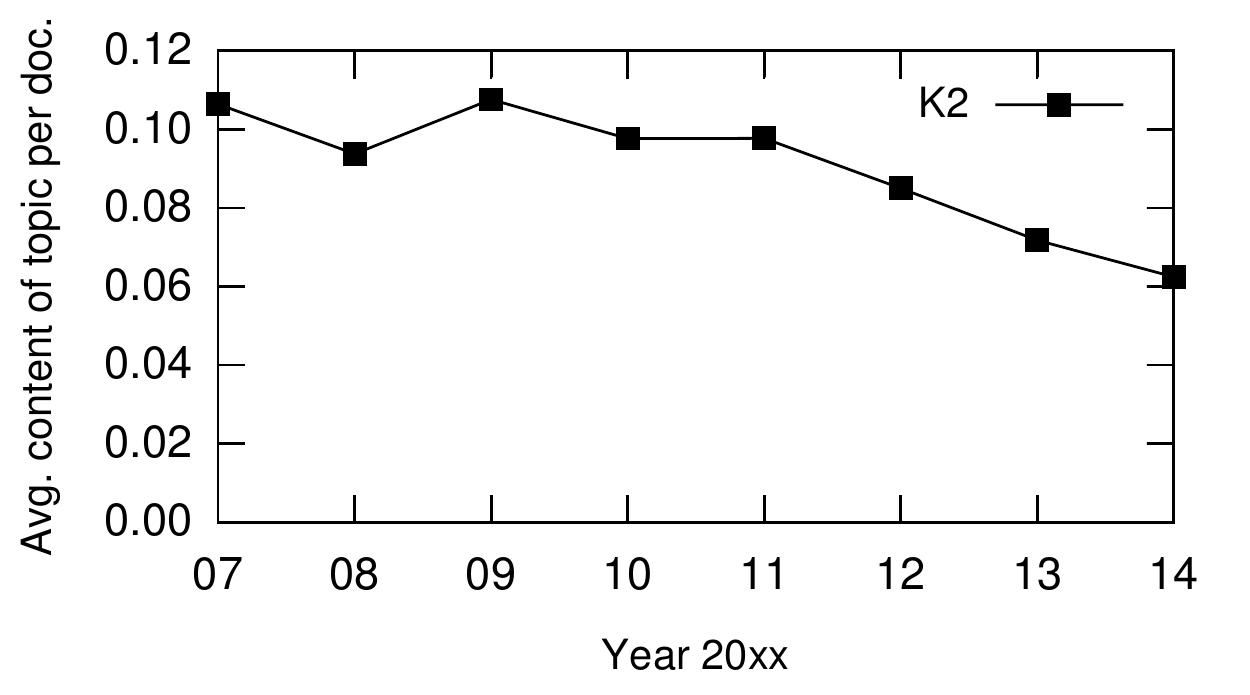}
\caption{Time trends for sample topics K1 (waste management)
         and K2 (construction works). \label{fig:TopicTrends}}
\end{figure}

Another interesting output of the LDA method is a percentage of every topic
in each document. This enables for analysis of the time trends
occurring in themes of appeals lodged in contract award procedures.  As an
example of such analysis, I extracted temporal behaviour for  topics
K1 (waste management) and K2  (construction works), see
Figure~\ref{fig:TopicTrends}.
The K1 theme has a very clear peak in 2013. This is definitely related to a
significant change of regulations in Poland, which forced all
municipalities to select new waste management company within tendering
procedure in this period.
The time trend for K2 clearly correlates with the financing perspective of
the Structural and Cohesion EU Funds (perspective 2007--2013) and
the 2012 UEFA Championships in Poland/Ukraine, which triggered a lot
of construction works.

I find the presented approach very promising.  It enabled for automatic
detection of many meaningful themes and provided interpretable time trends.
The results of LDA may be in future applied to improve information
retrieval and facilitate tasks such as clustering, searching or similar
documents finding.  Moreover, it could also form a valuable input material
for various analyses or reports prepared by legal experts.

\smallskip
{\bf Acknowledgments.}
I acknowledge the use of computing facilities of the ICM~HPC~Centre within
the grant G57-14. I acknowledge the support from the SAOS project financed
by the National Centre for Research and Development.

\bibliographystyle{unsrt}
\bibliography{biblio}
\end{document}